# Channel Equalization Using Multilayer Perceptron Networks

SABA BALOCH*, JAVED ALI BALOCH**, AND MUKHTIAR ALI UNAR***

## ABSTRACT

In most digital communication systems, bandwidth limited channel along with multipath propagation causes ISI (Inter Symbol Interference) to occur. This phenomenon causes distortion of the given transmitted symbol due to other transmitted symbols. With the help of equalization ISI can be reduced. This paper presents a solution to the ISI problem by performing blind equalization using ANN (Artificial Neural Networks). The simulated network is a multilayer feedforward Perceptron ANN, which has been trained by utilizing the error back-propagation algorithm. The weights of the network are updated in accordance with training of the network. This paper presents a very effective method for blind channel equalization, being more efficient than the pre-existing algorithms. The obtained results show a visible reduction in the noise content.

**Key Words:** Blind Channel Equalization, Neural Networks, Noisy Signal, Multi Layer Perceptron, Error-Back Propagation.

## 1. INTRODUCTION

With the passage of time, digital communication has almost prevailed analog communication. Prominent factors behind the current situation are the escalating demand and falling prices of digital equipment. Digital communication basically includes transferring of certain digital information for instance voice, images or data from the transmitting end to the receiving end, but the data transferred should be received in the actual form [1-2]. Practically this cannot be achieved. ISI is one of the most influential problems faced practically in digital communication. This causes distortion to some of the transmitted symbols due to other transmitted symbols. Performing equalization on the channel can minimize the ISI. The two major reasons of ISI in a channel are as follows:

(1) As the channel used for communication has a limited bandwidth, it causes the pulse waveform passing through it, to disperse or spread. If we consider a channel with a much larger bandwidth in comparison to the pulse bandwidth, the spread or dispersing of the pulse should be minimal. On the other hand when the bandwidth of the channel is almost same as the signal bandwidth, the spreading will exceed the symbol duration and cause the signal pulses to overlap [2-4]. This overlapping of symbols is called interference between symbols.

(2) Multipath is a signal propagation phenomenon due to which signals may reach the receiving antenna by two or more paths. This causes the transmitted signal to be dispersed in time, which results in overlapping of different transmitted symbols. This is also known as ISI, which can cause high error rates, if not compensated [2,4].

\*        Assistant Professor, Department of Electronic Engineering, Mehran University of Engineering & Technology, Jamshoro.
\*\*       Assistant Professor, Department of Computer Systems Engineering, Mehran University of Engineering & Technology, Jamshoro.
\*\*\*      Meritorious Professor, Department of Computer Systems Engineering, Mehran University of Engineering & Technology, Jamshoro

The ISI problem can be solved by devising a means to offset or minimize the ISI at the receiving end before detection. An equalizer can be used as a compensator for the ISI. Many equalization techniques have been proposed and implemented. In some techniques, there is a need to transmit a training sequence prior to signal transmission and some perform equalization without using a training sequence [5-6]. Studying the previous techniques showed the presence of noise even after the equalization process. This motivated us to propose a method which would reduce the noise to a minimal level. This can be achieved using ANNs, which has the advantage of accuracy and provides us with faster response. In the following section we have discussed channel equalization and its types.

## 2. CHANNEL EQUALIZATION AND BLIND EQUALIZATION

One of the most prominent functions for the receivers in many data communication systems is channel equalization. The requirement for data communication is that a specific analog medium be used to transmit the digital signals from the source to the receiver. Practical restraints in analog channels make them imperfect and may cause undesired distortions to be introduced [7-8]. In linearly distorted channels, the distortion can be effectively removed and compensated with the help of channel equalization. In other words inter symbol interference can be mitigated by performing equalization.

Here the equalizer coefficients are initially adjusted by transmitting a known trail sequence to the receiver. However, in certain situations sending a training sequence is either not feasible or is costly. A blind equalizer attempts to recover the ISI severed transmitted signal without using a training sequence. This can be achieved by computing the inverse of the channel. Under such circumstances it is required that the receiver synchronizes itself with the received signal and adjust the equalizer accordingly. This is known as blind equalization. The input signal and the channel characteristics define the performance of a blind equalizer [8-9]. The probability distribution for the channel input should also be known. The general function of a blind equalizer can be understood by Fig. 1.

Some good examples of blind equalization are cable modem and digital cable TV. The blind equalization technique makes use of the transmitted sequence statistics. Given are some basic techniques / algorithms used for performing blind equalization.

## 2.1 Least Mean Square Algorithm

This is a linear adaptive filtering algorithm that is based on the stochastic gradient algorithms [2]. Stochastic gradient defines a cost function based on mean of the squared error [10-11]. Then the steepest descent is computed by considering the minimal error on the error surface. This algorithm is made up of two parts. In the first half, the transversal filter output is calculated using the tap inputs and by computing difference between output of the filter and the desired/required response, an error term. In the second half, using the error term the tap weights are adjusted accordingly.

## 2.2 Constant Modulus Algorithm

CMA (Constant Modulus Algorithm) is one of the majorly used adaptive algorithms for performing blind channel equalization. In this technique, the constant modularity of the applied signal is used as the required property. Here any transmitted sequence which presents a constant phase offset can be considered as the right sequence at the receiver since the phase shift does not change the constant modularity property of a signal. This contrary to LMS (Least Mean Square) error surface gives us multiple minima. Out of which the most acceptable solution will be considered as the global minima and all other will be local minima. Due to this fact, the CMA has a slower convergence rate than LMS [2,10]. CMA provides low computational complexity. It is robust against additive noise and it also provides absence of cost dependent local minima. These are the reasons why CMA is preferred over other algorithms.

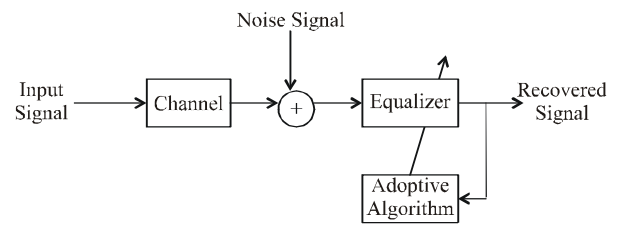

*FIG. 1. BLIND EQUALIZATION GENERAL BLOCK DIAGRAM*

## 2.3 Fractionally Space Constant Modulus Algorithm

By using fractionally spaced equalizers, we can develop another variation of transverse linear filters [5,12]. In FSE-CMA (Fractionally Space Constant Modulus Algorithm) the spacing interval between the equalizer taps is a small part of the total timing of the symbol. It samples the received signal at a rate higher than the input signal. This factor makes fractionally spaced equalizers to produce error in terms of time. Performance wise, FSEs are superior to other transverse equalizers [12].

## 3. ARTIFICIAL NEURAL NETWORKS

ANN is basically a computational system, which is based on the structure, the ability to learn and processing method just like the human brain. An ANN basically comprises of a huge amount of very simple elements inspired by neurons, along with a huge number of weighted connections between these processing elements [11]. Having distributed representation of knowledge over the connections, ANN acquires knowledge through a defined learning process.

The distinct features of ANNs include
- Huge parallelism
- Distributive representation
- Ability to learn
- Ability to generalize
- Fault tolerance

There are three fundamental elements of ANNs; Processing Units, topology and learning algorithm [11].

## 3.1 Perceptron/Multilayer Perceptron Network

The perceptron model comprises of only one neuron having a linear weighted net function and a threshold activation function. Every neuron in a given layer that is arranged in such a fashion is called as a perceptron [11,13]. Input will be in parallel to all the neurons in such a layer simultaneously. Perceptron networks having only one layer are able to classify linearly separable problems only. In situations where we have non-separable problems, one layer is not enough, and thus, it is required to use more layers. Network where along with the output layer one or more hidden layers that are made up of hidden neurons are introduced, are known as Multi-layer (feed-forward) network, as shown in Fig. 2.

In such a network each layer of neurons (i.e. each perceptron) is capable to dividing a space linearly. Considering this division process in two dimensions would be like drawing a straight line across the Cartesian grid and like slicing a cube into halves along any arbitrary plain in three dimensions. A similar partition can be considered for higher dimensions but that cannot be visualized. However, we can consider multiple cascaded layers, each layer performing a number of such processes, i.e. each neuron linearly partitioning a plane, but this partitioning must be along a different (hyper) plane for each layer. Considering one set of lines on the grid will give us binary partitioning: 0 or 1 [11,14] however, if we further partition the already partitioned space, it will further refine our specific region. Again if we consider that region and linearly divide it, we will obtain an even more refined region. And so on. Informally we can even show that any region can be defined in n-space by just using three layers [11]. Here the first layer will draw the lines. These lines will then be combined into convex hulls by the second layer. The third layer combines the convex hulls and forms arbitrary regions. In this way we can build a three layer cascaded perceptron NN known as MLP (Multilayer Perceptron). The MLPs were put into practice only when learning algorithms were developed for them, one of them being the error back propagation algorithm [11,14].

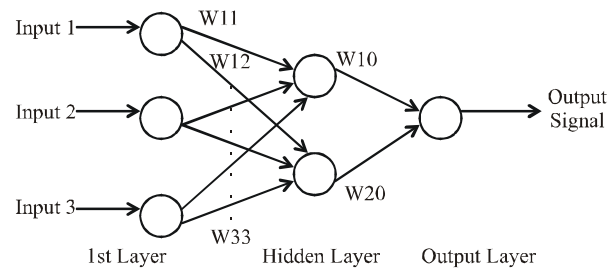

FIG. 2. A FULLY CONNECTED MLP NETWORK

### 3.2 Error Back-Propagation Algorithm

One of the most popular learning paradigms for MLP networks is the Error back-propagation algorithm [11]. This algorithm uses the squared error measure for output nodes, i.e. it works on the principle of the delta rule. Consider a perceptron weight $w_{ji}$; generalized delta rule is used to update the weight corresponding to a connection from neuron i to neuron j. There are two phases of the generalized delta rule. During the first phase the output values for each unit are computed for the applied inputs. The computed output value is then compared with the required value. This comparison gives an error term for each output unit. During the second phase a reverse pass takes place throughout the network. This involves the error term proceeding towards each unit within network and the required weight alterations are computed accordingly [10-11,14].

$$w_{ji} = w_{ji} + \Delta w_{ji} \quad (1)$$

$$\Delta w_{ji} = -\eta \left( \frac{\partial \varepsilon}{\partial w_{ji}} \right) \quad (2)$$

Where $\varepsilon$ shows the instantaneous sum of squared errors and $\eta$ is the learning rate. This learning rate will decide the speed in which the weights will be able to adjust for every time advance. The negative sign shows that the changing of weights will reduce the error [11].

### 4. NETWORK DESIGN OF MLP BASED BLIND EQUALIZER

In this paper we have designed an MLP network architecture. The network is a 3-layer feedforward network. The network comprises of nine neurons; eight neurons are used within the hidden layer and one neuron is used in the output layer. Tangent sigmoid activation function is used for the hidden layer and linear activation function for the output layer [7,10-11].

Error back-propagation learning algorithm is used to train the network. The network weights and biases are automatically adjusted with training process. The simulations were performed in MATLAB and its Neural Network Toolbox.

### 4.1 Simulation and Results

The basic goal of the designed network is to perform blind equalization by using Multi-Layer Perceptron network and then comparing the obtained results with the previously implemented algorithms designed for conventional blind channel equalization. MATLAB was used to perform the simulations on M-QAM (M-Ary Quadrature Amplitude Modulation) modulated signals [15]. The input signal is subjected to white Gaussian noise. LMS Algorithm, CMA and FSE-CMA were also simulated [5,9]. Fig. 3 shows the 4-QAM constellation diagram of our transmitted signal and Fig. 4 shows the transmitted signal after being subjected to noise which, in our case is AWGN (Adaptive White Gaussian Noise). Figs. 5-7 show the received signal after being equalized using LMS, CMA and FSE-CMA blind equalization methods respectively [5,9]. As it can be seen that the transmitted symbols have been recovered but the presence of noise is still visible. Fig. 8 show the training of our MLP based equalizer. Here the goal was set to 0.0001 and our network achieved it in 366 epochs. Fig. 9 shows the recovered signal using our MLP network and it clearly has the least noise content than the previously developed algorithms.

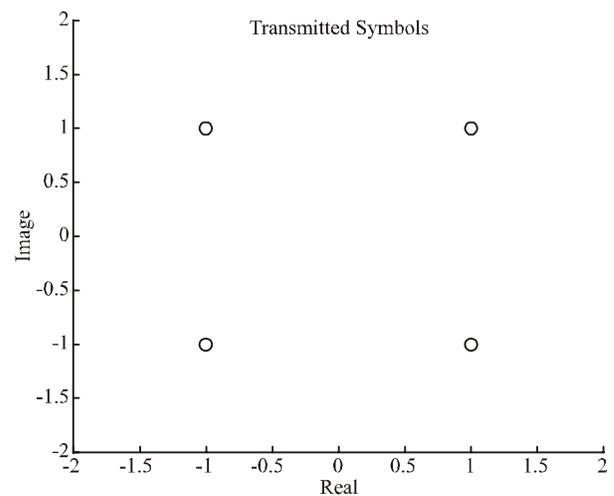

*FIG. 3. CONSTELLATION DIAGRAM OF 4 QAM SIGNAL*

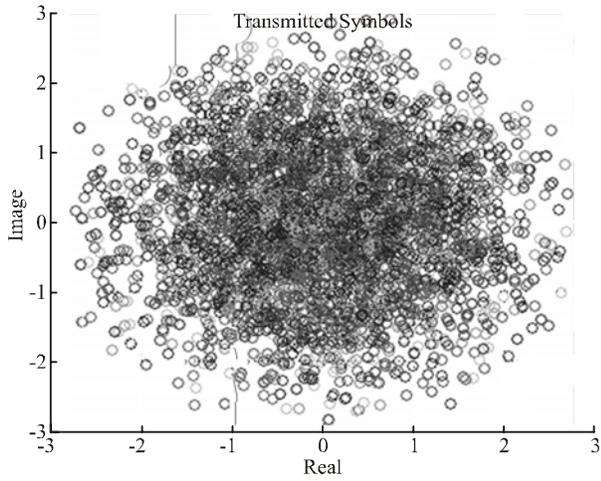

FIG. 4. RECEIVED 4 QAM NOISY SYMBOLS

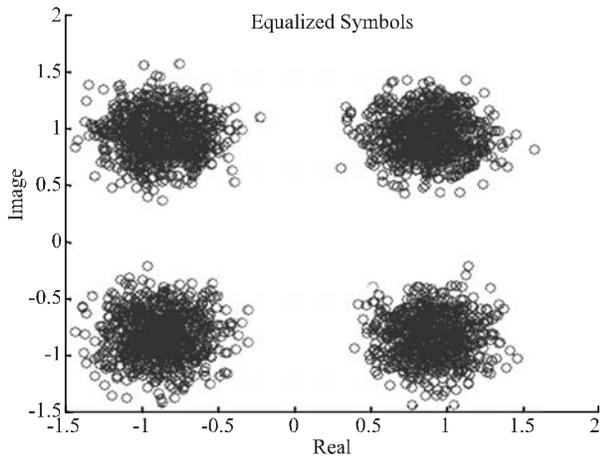

FIG. 5. 4-QAM EQUALIZED SYMBOLS USING LMS ALGORITHM

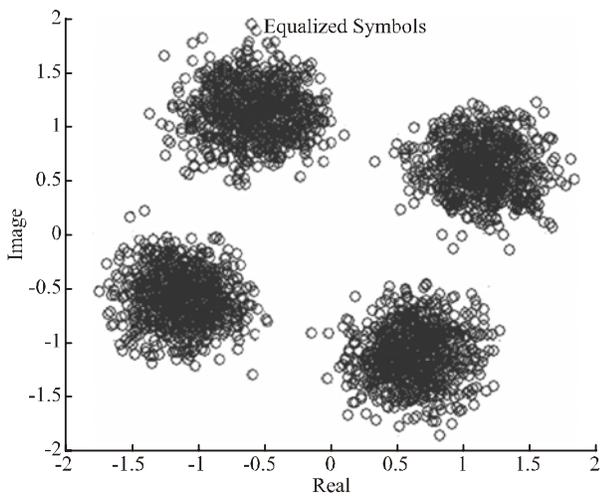

FIG. 6. EQUALIZED 4-QAM SYMBOLS USING CMA ALGORITHM

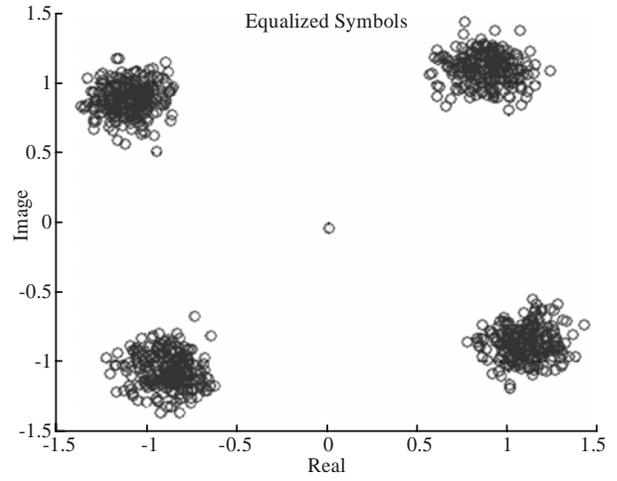

FIG. 7. EQUALIZED 4-QAM SYMBOLS USING FSE-CMA

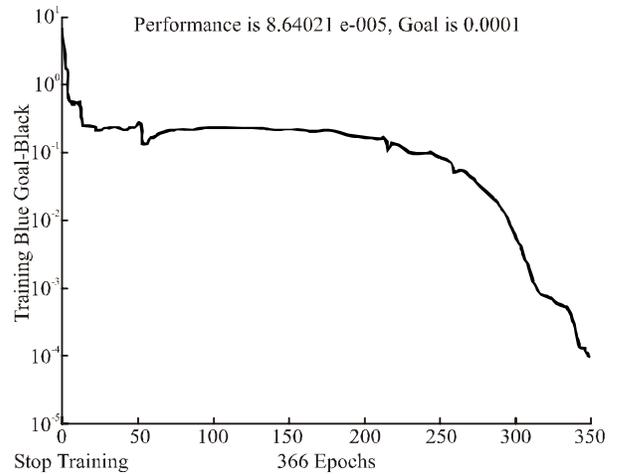

FIG. 8. TRAINING OF THE MLP BASED NEURAL NETWORK EQUALIZER-ACHIEVING GOAL IN 366 EPOCHS

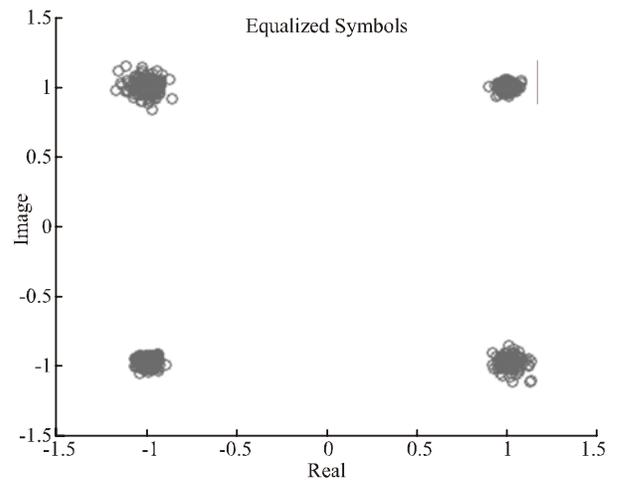

FIG. 9. EQUALIZED 4-QAM SYMBOLS USING MLP NETWORK, NEURAL NETWORKS

# 5. CONCLUSIONS

In this paper blind equalization using ANNs was performed. This was achieved by using the neural networks toolbox in MATLAB. By using MATLAB communication toolbox, noisy and distorted signals were generated by subjecting the randomly generated QAM or Q-PSK modulated symbols with AWGN. The ANN based on MLP was trained using the error back-propagation algorithm. The same noisy data was then applied to some of the most common blind equalization methods (LMS, CMA and FSE-CMA). The ANN, MLP network showed much better results than the above mentioned methods. According to the previous work done, some algorithms converged fast and responded quicker but lacked in terms of fully eliminating the noise term from the received signals and some algorithms mitigated the noise signal present but converged or responded slower. By using ANN we are able to achieve both advantages, i.e. the network training time was very less and according to the results our network showed minimum noise in the received signal. Our network is simpler in terms of calculations and computations and apart from requiring fewer taps, it takes lesser time in decoding the channel inputs than CMA blind equalizers. This shows that MLP based blind equalizer proves to be a good alternate for channels with severe ISI.

This research can be further extended by considering higher bit rates such as 8-QAM or even 16-QAM. Different ANN models and learning methods can be implemented along with varying the number of neurons and hidden layers.

## ACKNOWLEDGEMENTS

The authors would like to thank Prof. Dr. Abdul Karim Baloch, and Prof. Dr. Bhawani Shankar Chowdhry, Institute of Information & Communication Technologies, Mehran University of Engineering & Technology, Jamshoro, Pakistan, for their valuable suggestions and support.

## REFERENCES

[1] Moonen, M., and Proudler, I., "An Introduction to Adaptive Signal Processing", Kardinaal Mercierlaan, Belgium, 1999.

[2] Khatri, R., "Blind Channel Equalization Algorithms", B.E. Thesis, Mehran University of Engineering & Technology, Jamshoro, Pakistan, 2006.

[3] Haykin, S., "Adaptive Filter Theory", Prentice Hall, 2005.

[4] Sklar, B., "Digital Communications", Pearson Education, 2006.

[5] Li, X., "Channel Estimation/Equalization in Digital Communications", Tutorial on Channel Equalization, Binghamton University, New York.

[6] Wong, T.F., and Lok, T.M., "Theory of Digital Communication", University of Florida, 2004.

[7] Paul, J., and Vladimirova, T., "Blind Equalization with Recurrent Neural Networks Using Natural Gradient", 3rdInternational Symposium on Communications, Control and Signal Processing, Malta, 12-14 March, 2008.

[8] Ding, Z., and Li, Y., "Blind Equalization and Identification", Marcel Dekker, Inc. 2001.

[9] Jamali, A.G., "Performance Comparison of Blind Channel Equalization Methods", M.E. Thesis, Mehran University of Engineering & Technology, Jamshoro, Pakistan, 2006.

[10] Lee, C.M.,Yang, S.S., and Ho, C.L., "Modified Back-Propagation Algorithm Applied to Decision-Feedback Equalization", IEE Proceedings of Vis. Image Signal Process, Volume 153, No. 6, December, 2006.

[11] Haykin, S., "Neural Networks - A Comprehensive Foundation", McMillan, 2007.

[12] Gong, K., Dong, Z., and Ge, L., "A Joint Algorithm Between Timing and Fractionally-Spaced Equalization Applied to QAM Signal", International Conference on Electronics, Communication & Control, Zhenjiang, 03 November, 2011.

[13] Wong, C., and Fine, T., "Adaptive Blind Equalization Using Artificial Neural Networks", Cornel University, New York, 1996.

[14] Tebelskis, J., "Speech Recognition using Neural Networks", Ph.D. Thesis, School of Computer Science, Carnegie Mellon University, Pittsburgh, Pennsylvania, 1995.

[15] www.mathworks.com